%% file: elsarticle-template-1-num.tex
\documentclass[preprint,12pt]{elsarticle}

\usepackage{graphicx}
\usepackage{amssymb}
\usepackage{hyperref}
\usepackage{placeins}
\usepackage{lineno}

\journal{arxiv.org}

\begin{document}

\begin{frontmatter}


\title{Choosing the Right Word: Using  Bidirectional LSTM Tagger for Writing Support Systems}



\author{Victor Makarenkov}

\author{Lior Rokach}

\author{Bracha Shapira}

\address{Department of Software and Information Systems Engineering \\
Ben-Gurion University of the Negev\\
Beer-Sheva, Israel}

\begin{abstract}
Scientific writing is difficult. It is even harder for those for whom English is a second language (ESL learners). Scholars around the world spend a significant amount of time and resources proofreading their work before submitting it for review or publication.

In this paper we present a novel machine learning based application for proper word choice task. Proper word choice is a generalization the lexical substitution (LS) and grammatical error correction (GEC) tasks. We demonstrate and evaluate the usefulness of applying bidirectional Long Short Term Memory (LSTM) tagger, for this task. While state-of-the-art grammatical error correction uses error-specific classifiers and machine translation methods, we demonstrate an unsupervised method that is based solely on a high quality text corpus and does not require manually annotated data. We use a bidirectional Recurrent Neural Network (RNN) with LSTM for learning the proper word choice based on a word's sentential context. We demonstrate and evaluate our application on both a domain-specific (scientific), writing task and a general-purpose writing task.
We show that our domain-specific and general-purpose models outperform state-of-the-art general context learning.
As an additional contribution of this research, we also share our code, pre-trained models, and a new ESL learner test set with the research community.
\end{abstract}


\end{frontmatter}

\input{1_Introduction}

\input{2_Background}

\input{3_Method}
\input{4_Evaluation}
\input{5_Results}
\input{6_Conclusions}
\section {ACKNOWLEDGEMENTS}

We would like to express our gratitude to Stas Shteibook and Evgenyi Vainer for their invaluable technical support in setting up the GPU machine and compiling the DyNet toolkit in the Windows environment. We also thank Roee Aharoni for numerous comments and helpful advice during the experiments that are described in this work. We would also like to express our gratitude to the two editors, who contributed to this research: Robin Levy-Stevenson and Jade Makarenkov.

\input{7_Appendix}



\newpage
\newpage

\bibliographystyle{model1-num-names}
\bibliography{sample.bib}







\end{document}

%% file: 1_Introduction.tex
\section{Introduction}
Writing good scientific papers is a difficult process, and this task is even more challenging for non-native English speakers. In this paper we propose a new task of  \textit{proper word choice}. We apply a bidirectional LSTM tagger \cite{2015arXiv151100215W} to implement a writing support method for those for whom English is a second language (referred to as ESL learners in this paper). Our application uses long term linguistic patterns and regularities which are captured by RNNs with an LSTM cell \cite{Hochreiter:1997:LSM:1246443.1246450}, when extracted from a large scientific corpus. This application could be used to help ESL learners with tasks such as choosing the right word, avoiding homophone confusion, and using correct and up-to-date jargon and terminology.
Our application is not simply another spell-checker; rather it is an implementation of a method specially aimed at assisting ESL learners and is focused on suggesting corrections for the types of errors made by these writers. 

There are hundreds of millions of non-native English speakers worldwide, yet most prestigious academic conferences are held in the English language. A paper's acceptance or rejection often hinges on the quality of writing, which is why scholars invest so much time and resources in writing, editing, spell-checking, proofreading, and revising their work before submitting it for review.

In order to meet the needs of ESL learners, we address two important word replacement tasks, each of which are generalized: the grammatical error correction (GEC) and lexical substitution (LS) tasks. In the GEC task, the need to replace one word with another derives from the inherent definition of the \textit{error correction} task, in which a written sentence does not comply with English grammar rules; consequently, a word replacement can inadvertently change the entire meaning of the sentence. In contrast, in the LS task, the substitute for a word in a sentential context, must preserve the original meaning \cite{mccarthy2009english} of the sentence, while the sentence itself must adhere to English grammar rules. Both reasons for word replacement (because of a grammatical error or a poor word choice) occur frequently in the articles of ESL learners. 

The main contributions of this work are threefold: 1) We define a new task of proper word choice and develop a solution for it. This task is a generalizartion of GEC and LS. 2) We present an RNN based model, that addresses many various types of GEC and the LS in a \textit{single} model (rather than an ensemble).  3) While most exsiting solutions based on supervised learning, our model relies solely on an \textit{unsupervised} data, without any human annotation, that results in a comprehensive, unsupervised, and powerful system for ESL writing assistance. In order to better demonstrate our  application's usefulness, we created an ESL learner test-set. We share our code, pre-trained models, and the ESL learner test-set for the research community.

%% file: 2_Background.tex
\section{Background and Related Work}
The proposed task of proper word choice is closely related to other error correction tasks, including: semantic collocation correction \cite{Dahlmeier:2011:CSC:2145432.2145445}, lexical substitution \cite{McCarthy:2007:STE:1621474.1621483}, paraphrase generation \cite{madnani2010generating}, grammatical error correction \cite{ng-EtAl:2014:W14-17} and sentence completion\footnote{Microsoft Sentence Completion Challenge} \cite{Zweig2011}. In this section we present background and related work associated with the GEC and LS tasks which are generalized in our task. This section also addresses the related topics of bidirectional LSTM tagger and language modeling.

\begin{table*}[]
\centering
\caption{Examples of common grammatical errors and error types in the CoNLL-2014 shared task. The errors are marked with an asterisk (*) and the correction follows the forward slash (/). Examples are provided from \cite{ng-EtAl:2014:W14-17}.}
\begin{tabular}{|l|l|}
\hline
\multicolumn{1}{|c|}{\textbf{Error type}} & \multicolumn{1}{c|}{\textbf{Example}} \\ \hline
Verb tense & \begin{tabular}[c]{@{}l@{}}Medical technology during this time \textit{*is/was}\\ not advanced enough to cure him.\end{tabular} \\ \hline
Verb modal & \begin{tabular}[c]{@{}l@{}}Although the problem \textit{*would/may} not be serious, \\ people \textit{*would/might} still be afraid.\end{tabular} \\ \hline
Verb form & \begin{tabular}[c]{@{}l@{}}A study in 2010 \textit{*shown/showed} that patients \\ recover faster when surrounded by family members.\end{tabular} \\ \hline
Subject-verb agreement & \begin{tabular}[c]{@{}l@{}}The benefits of disclosing genetic risk information\\  \textit{*outweighs/outweigh} the cost.\end{tabular} \\ \hline
Article or determiner & \begin{tabular}[c]{@{}l@{}}It is obvious to see that \textit{*internet/the internet} saves\\ people time and also connects people globally.\end{tabular} \\ \hline
Noun number & \begin{tabular}[c]{@{}l@{}}A carrier may consider not having any \\ \textit{*child/children} after getting married.\end{tabular} \\ \hline
Preposition & \begin{tabular}[c]{@{}l@{}}This essay will \textit{*discuss about/discuss} whether the \\ carrier should tell his relatives or not.\end{tabular} \\ \hline
Wrong collocation & \begin{tabular}[c]{@{}l@{}}Early examination is \textit{*healthy/advisable} and will \\ cast away unwanted doubts.\end{tabular} \\ \hline
Word form & \begin{tabular}[c]{@{}l@{}}The sense of \textit{*guilty/guilt} can be more than expected.\end{tabular} \\ \hline
\end{tabular}

\label{table:connexamples}
\end{table*}

\subsection{Grammatical Error Correction}
In recent years, several GEC competitions have taken place, including the HOO-2011 \cite{dale2011helping}, HOO-2012 \cite{dale2012hoo}, CoNLL-2013 \cite{Ng_theconll-2013}, and the CoNLL-2014 \cite{ng-EtAl:2014:W14-17} competitions. The CoNLL-2014 shared task on GEC, the most recent and prominent of these competitions, proposed 28 error types, the most common of which are presented in Table \ref{table:connexamples}. In this competition, the participants were required to develop end-to-end GEC systems, for specific, non-exhaustive, grammatical error detection and correction. The participating teams developed a wide range of approaches, including statistical classifiers \cite{DBLP:conf/conll/RozovskayaCSRH14}, language models, and statistical machine translation \cite{grundkiewicz2014amu}, and rule-based modules \cite{felice2014grammatical}. The systems submitted were trained on an NUCLE \cite{dahlmeier2013building} annotated corpus of articles written by ESL learners. More recent GEC research has used machine translation and classifiers \cite{mizumoto2016discriminative, susanto2015systems}, and the state-of-the-art GEC method was developed by Rozovskaya and Roth \cite{rozovskaya2016grammatical} who combined machine translation methods and machine learning classifiers in the current best performing method using the CoNLL-2014 test set.

\subsection{Lexical Substitution}
The LS task \cite{mccarthy2009english} has attracted increased attention following its inclusion in SemEval-2007 \cite{McCarthy:2007:STE:1621474.1621483}. In this type of word replacement, given a word in a sentence, the task is to substitute the correct word, such that the original meaning of a sentence remains the same. 
For example \cite{mccarthy2009english}:
\begin{itemize}

\item "After the \{ match / game \}, replace any remaining fluid deficit to prevent problems of chronic dehydration throughout the tournament."

\item "The results clearly \{ indicate / show / illustrate \} that our method outperforms the current state-of-the-art."

\end{itemize}
The word can be one of the following four part-of-speech (POS) classes: noun, verb, adjective, or adverb. The systems presented at Sem-Eval 2007 relied heavily on multiple external resources and inventories such as WordNet \cite{Miller:1995:WLD:219717.219748}. The LS task involves the subtasks of candidate prediction and candidate ranking, and most recent research has only focused only on candidate ranking. However, Melamud et al. \cite{DBLP:conf/conll/MelamudGD16} used bidirectional LSTM for context learning, and this represents the current state-of-the-art approach in LS.

\subsection{Bidirectional LSTM tagger}

\begin{figure*}[]
    \centering
    \includegraphics[width=0.60\textwidth]{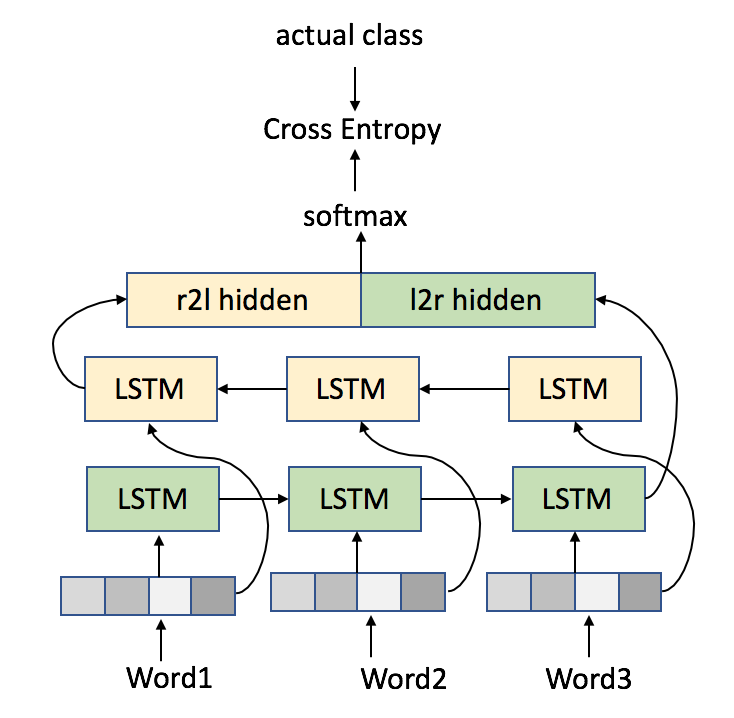}
    \caption{An example of bidirectional LSTM tagger architecture.}
    \label{fig:bilstm_tagger}
\end{figure*}

Figure \ref{fig:bilstm_tagger} presents the approach of bidirectional LSTM tagger when applied to  a text classification task. Each word is transformed into its dense embedding representation and is then being fed into two LSTM networks: 1) left-to-right LSTM network, and 2) right-to-left LSTM network. The outputs of the two networks (the LSTM hidden states) are further concatenated and the classification is performed. When applying this approach to text, a model learns from target word's prefix (left-to-right) and suffix (right-to-left) contexts and performs the final classification based on the jointly learned representation of this context.
Similar bidirectional LSTM models were successfully used in various NLP tasks, e.g., Part-of-Speech (POS) tagging  \cite{P16-2067}, context learning \cite{DBLP:conf/conll/MelamudGD16} and sentiment analysis \cite{ISI:000392770900019}. 
We formally describe the application of bidirectional LSTM tagger to the proper word choice task in section \ref{sec:formal-model}.

\subsection{Deep Language Modeling}
A language model (LM) is defined as a function that assigns a probability distribution over words in a given vocabulary \cite{IntroIR2008}.  While n-gram based LMs have been studied for decades, powerful deep learning and RNN-based LMs (RNNLM) have only begun to attract a significant amount of attention in recent years \cite{zaremba2014recurrent, jozefowicz2016exploring, shazeer2017outrageously, DBLP:journals/corr/MerityXBS16}, along with other text classification tasks, such as sentiment analysis \cite{ISI:000379634700032, ISI:000392770900019, ISI:000397687400019, ISI:000414619600010} or fake-news detection \cite{Ma:2016:DRM:3061053.3061153,DBLP:journals/corr/abs-1708-01967,DBLP:journals/corr/RuchanskySL17}. Training an LM with an RNN as its sequence model paradigm enables models to be created that reflect a target word's long and varied history.

%% file: 3_Method.tex
\section{Bidirectional LSTM tagger for proper word choice}

We now turn to formally specify the proper word choice task, elaborate on the application of bidirectional LSTM tagger to the task, and formally define it as well.

\subsection{Formal task specification}
We formally define the task as following:\\
Given a sentence $s=<w_1, w_2,...,target,...,w_n>$ where the $target$ is explicitly specified, the task is to provide a sorted list of the $k$ most appropriate words to replace the $target$ in $s$ based on the sentential context of the $target$ in $s$. 

\begin{figure*}[]
    \centering
    \includegraphics[width=1.0\textwidth]{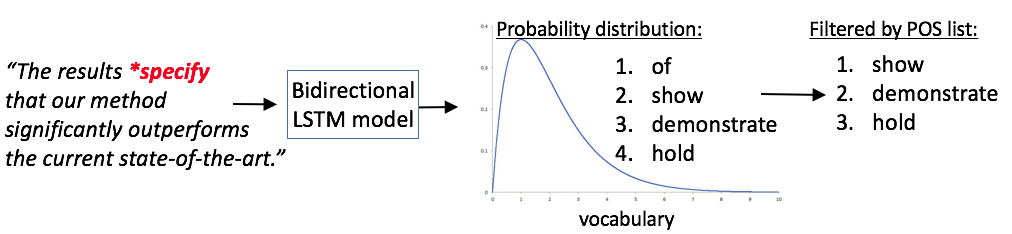}
    \caption{Overview of the proposed bidirectional LSTM application for word choice method. For example, in a given sentence, for the target word, \textit{specify}, the method suggests a list of three replacement candidates, sorted by the probability of observing a candidate in this sentential context.}
    \label{fig:method}
\end{figure*}

\subsection{Application overview}

Our main goal is to predict the most appropriate target word, given the original word written by an ESL writer and its sentential context.
The application consists of the following steps which are presented in Figure \ref{fig:method}:
\begin{enumerate}
\item Obtain a probability distribution from the bidirectional LSTM model, which was trained to predict a target word based on its sentential context.
\item From this distribution, filter the words that own  such a sense that has the same POS as the original target word that was written  by the ESL writer. The sense list and respective POS tags are extracted from WordNet \cite{Miller:1995:WLD:219717.219748}. 
\end{enumerate}

\begin{figure*}[]
    \centering
    \includegraphics[width=1.0\textwidth]{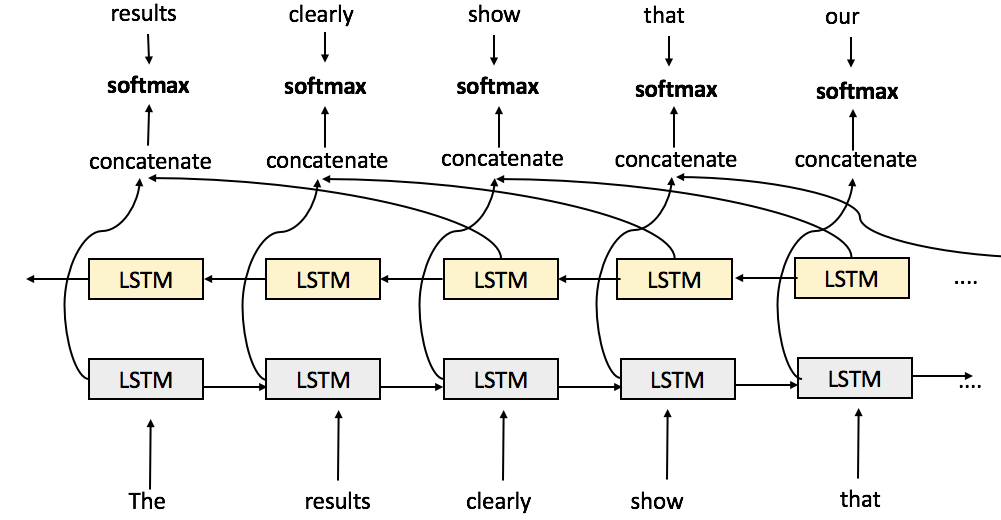}
    \caption{An application of bidirectional LSTM tagger to the proper word choice task. For example, for learning to suggest the word, \textit{results}, we concatenate the left-LSTM network's output before the word, \textit{the}, and right-LSTM before same word from the right side, \textit{clearly}.}
    \label{fig:model}
\end{figure*}

The bidirectional LSTM tagger approach lies at the heart of our application. In essence, our model combines an RNN approach to language modeling and variable-length context learning around the target word. The model's architecture is presented in Figure \ref{fig:model}. The model consists of two separate recurrent neural networks with LSTM memory cells. To train this model we feed the sentence (from left to right) into the first LSTM, and then we feed the same sentence (right to left) into the second LSTM. The outputs of the two networks are concatenated and represent the sentential context embedding as illustrated in Figure \ref{fig:model}. The concatenation of the networks' outputs does not occur as expected from the standard bidirectional LSTM tagger, in which the outputs at time $t$ are combined for each LSTM with its relative bidirectional zipped counterpart. Instead, the outputs of two LSTM networks are combined with an offset of two as displayed in Figure \ref{fig:model}. Finally, the combined output undergoes \textit{softmax} normalization, in order to obtain a probability distribution on the words in the vocabulary. We use the \textit{cross-entropy} as the \textit{loss} function in the process of learning our model's parameters. In this learning process we also learn the LSTM networks' parameters, the sentential context embeddings, and the target words' embeddings.

Consider the following sentence which was written by an ESL learner: \textit{The results clearly *indicate/show that our method outperforms the current state-of-the-art methods.} The word, \textit{indicate}, was originally written, but was later replaced with the word \textit{show} for a better word choice. In order for our method to produce such a suggestion, the word, \textit{show}, had to be learned by a model to appear in this sentential context.

\subsection{Formal application specification}
\label{sec:formal-model}
Using the notation of \textit{context2vec} \cite{DBLP:conf/conll/MelamudGD16}, let $lLS$ be the LSTM which reads the words from left to right, and let $rLS$ be the LSTM which reads the words from right to left. Let $w_{1:n}$ be a sentence of $n$ words. The context representation for the target word $w_i$ is defined as the concatenation of the outputs of $lLS$ and $rLS$:

\begin{equation}
biLS(w_{1:n},i)=lLS(l_{1,i-1}) \oplus rLS(r_{n:i+1})
\end{equation}

where $l$ and $r$ represent distinct left-to-right and right-to-left word embeddings of the words in the sentence. The concatenated representation is then linearly transformed to the vector of the vocabulary size\footnote{equal to 30,000 in this work}:

\begin{equation}
LIN(x) = W * x + b
\end{equation}
where $W$ is a matrix of model parameters, and $b$ is a vector representing the bias.
Next, we squash the $k$-dimensional vector $z=LIN(x)$ with a softmax function, to obtain a probability distribution:

\begin{equation}
\sigma(z)_i = \frac{e^{z_i}}{ \sum\limits_{j=1..k} e^{z_j}}
\quad\mathrm{for}\quad 
i=1..k
\end{equation}
 
Next, the loss is calculated using the cross-entropy function.
The distribution obtained is then used as an estimated distribution $Q$, and the true distribution $P$ assigns 1 to the target word and 0 to all of the other words:

\begin{equation}
H(P,Q) = - \sum\limits_{x}p(x)\log q(x)
\end{equation}

The LSTM equations that are computed at each step $t$, for each LSTM network are a implemented using coupled input and forget gates \cite{greff2016lstm}:
\begin{equation}
i_t = \sigma(W_{i}x_t + U_{i}h_{t-1} + Y_{i}c_{t-1} + b_i )
\end{equation}
\begin{equation}
f_t = \sigma(W_{f}x_t + U_{f}h_{t-1} + Y_{f}c_{t-1} + b_f )
\end{equation}
\begin{equation}
o_t = \sigma(W_{o}x_t + U_{o}h_{t-1} + Y_{o}c_{t-1} +b_o )
\end{equation}
\begin{equation}
g_t = \tanh(W_{g}x_t + U_{g}h_{t-1} +b_g )
\end{equation}
\begin{equation}
c_t = f_t \cdot c_{t-1} + i_t \cdot g_t
\end{equation}
\begin{equation}
h_t = o_t \cdot \tanh(c_t)
\end{equation}
where $W_{*}$ , $U_{*}$ and $Y_{*}$ are the LSTM model parameters, and $b_{*}$ are the biases. $i_t$, $f_t$, and $o_t$ are the input gate, forget gate, and output gate, respectively. $h_t$ is the output that is passed to the next layer of the network, and $c_t$ is the state that is passed to the next step at time $t+1$.

\begin{figure*}[]
    \centering
    \includegraphics[width=0.9\textwidth]{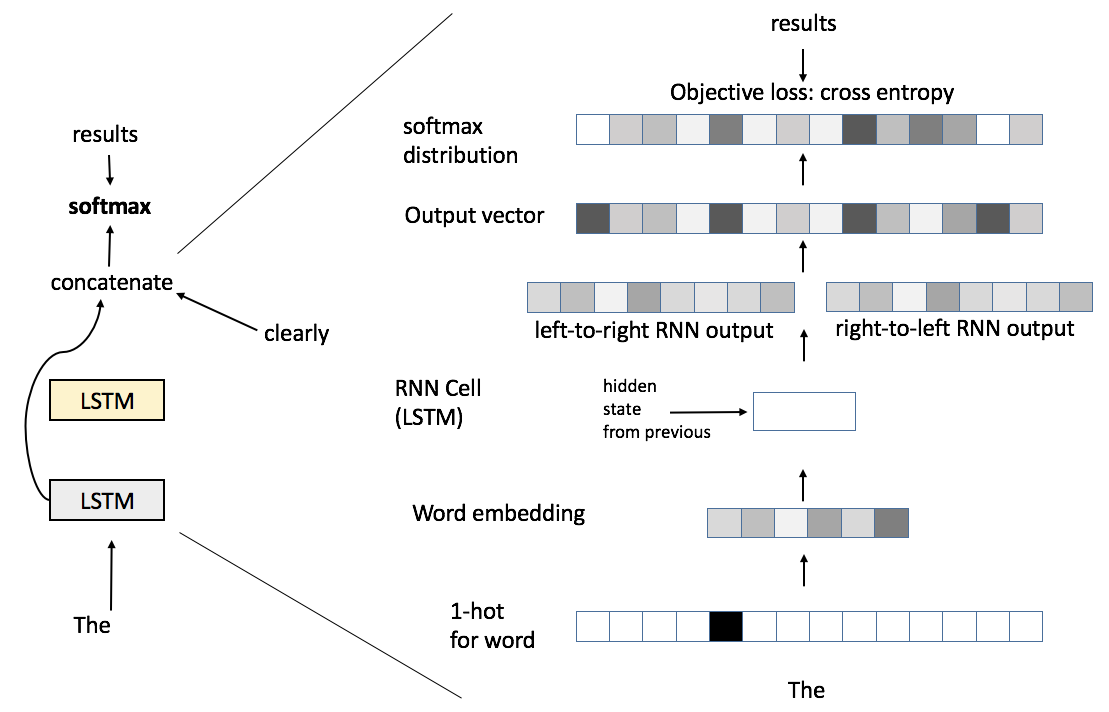}
    \caption{A detailed example of the training process. In each step, a target word (\textit{results} in this example) is learned based on a prediction given the two words immediately surrounding the target word, which in turn, are preceded with sentential context.}
    \label{fig:model_detailed}
\end{figure*}

We follow Aharoni et al. \cite{aharoni2016} in a way of demonstrating the step-by-step training. The graph in Figure \ref{fig:model_detailed} presents the training at each step for a given target word. For the left-to-right LSTM, the one-hot encoded context word is transformed to a learned ad hoc embedding. Then, the embedding is used, in turn, as an input to the LSTM memory cell, which also combines its previous state. Next, the hidden state of LSTM is concatenated with its respective counterpart from right-to-left LSTM which was learned separately in an analogous manner. The concatenated representation of the context is transformed to a distribution of vocabulary size and used to compute the loss cross-entropy function against the actually observed word in the training text.

\subsection{The Role of WordNet In POS Filtering}
A crucial part of our application is filtering only those words, for which the POS in the sentential context is identical to the POS of the originally written word. We experimented with two different POS extraction procedures: 1) the NLTK \cite{bird06nltk} POS tagger which was trained on the Penn Treebank \cite{Marcus:1993:BLA:972470.972475} corpus, and 2) the WordNet \cite{Miller:1995:WLD:219717.219748} electronic dictionary of the English language. 

In the case of NLTK, we put a candidate word in place of the originally written word and applied the POS tagging procedure to determine the candidate word's POS tag.

In the case of WordNet, we extracted all possible senses of a candidate word and the originally written word and checked whether both words share the same POS in one of their senses.

We found that our application achieved slightly better performance when using WordNet as the POS source for the candidate words, as we discuss in detail in Section \ref{section:results}.
Intuitively, using WordNet rather than a statistical POS tagger impairs the purity of the statistical method, because a hand-crafted word resource is used in the suggestion method. However, since this strategy resulted in a slightly higher mean reciprocal rank (MRR) value without affecting the results and final conclusion, we used it in our experiments.

%% file: 4_Evaluation.tex
\section{Evaluation Settings}
\label{section:evaluation}
We evaluate our application in three different experimental settings. First, we demonstrate how a piece of scientific writing (176 sentences) can be improved using our application with a model learned from a high quality scientific text corpus. Second, to further demonstrate the performance of our application, we apply it to general-purpose writing using a subset of the CoNLL-2014 shared task \cite{ng-EtAl:2014:W14-17} dataset as the test set (2421 sentences). Third, we perform a human evaluation, using professional English editors' judgments to evaluate our method's performance.  

Like the evaluation conducted in the CoNLL-2014 \cite{ng-EtAl:2014:W14-17} competition and query auto completion (QAC) \cite{cai2016learning,cai2016survey} evaluation practice, we use a strict evaluation policy in the two first evaluation settings; that is, we compare the application's performance against the single correct word replacement suggested by a professional English editor, in order to provide an underestimate of the possible MRR. The word choice method generates a list of the top $k$ replacement suggestions, based on the original word's POS and sentential context.  We report the MRR measure obtained during our experiments. We set $k=100$ in our experiments.

Several other tasks allow multiple replacements. For example, consider the LS task evaluation \cite{McCarthy:2007:STE:1621474.1621483} method. Evaluating our application against a dataset with multiple possible word suggestions dramatically improves the results. For the third evaluation setting, we show an even better result, which is achieved by actual MRR obtained from the human evaluation.

In addition, in a special evaluation setting, we evaluate our application on this article and present the results in Appendix A.

\subsection{Scientific Writing}
Motivated by the challenge of writing high quality academic articles for top tier scientific conferences and journals, we focused on the following setting.

\noindent \textbf{High quality scientific corpus.} In order to train an effective word suggestion model\footnote{This pre-trained model is available at: \url{ https://drive.google.com/file/d/0B5iAITPoL9L2ald0VzNxZWtxdGM/view?usp=sharing}}, we used a high quality corpus of scientific articles. This corpus consists of 29,848 articles ranked $A$ and $A^*$ from 60 top ranked ACM and IEEE conferences that were selected on the basis of the CORE Rankings Portal\footnote{The Computing Research and Education Association of Australasia, CORE}. To accelerate the training process we used the 30,000 most frequently appearing words in the corpus as the vocabulary and only used sentences that contained 40 words or less. The resulting corpus consists of roughly 40 million tokens in 2.78 million sentences. The initial corpus size was 41 million tokens in 2.8 millions sentences, so the reduction was actually quite negligible.

\begin{figure}[t!]
    \centering
    \includegraphics[width=1.0\textwidth]{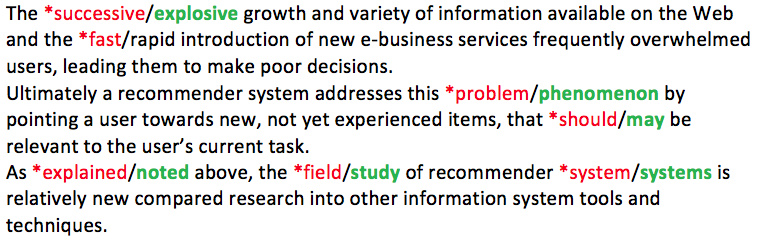}
    \caption{A snapshot of some of the corrections suggested for scientific articles written by ESL learners. Three different sentences after they were edited by a professional English editor. In the course of this research we collected these examples and constructed the new ESL learner test set.}
    \label{fig:dataset}
\end{figure}

\noindent \textbf{Scientific text test set.} To create a scientific text test set, we manually collected 176 sentences from 10 scientific articles written by researchers for whom English is a second language (ESL learners), including research students and faculty members from three different universities: Queen Mary University of London, Ben-Gurion University of the Negev and Tel Aviv University. The mother tongue of the writers was either Russian or Hebrew, which contributes to the diversity of the test set used in our experiment; the test set of the CoNLL-2014 shared task was derived from articles written at the National University of Singapore. Each of the 10 articles was written in the course of the scholar's ongoing research work, either as Ph.D. thesis chapters or articles for scientific publication (for a conference or journal). Editing of the papers was performed by four professional English editors, prior to the initiation of the experiments described in this work, so the editors did not know that their corrections would be used in this way. We selected corrections in which a single word was replaced by another single word. We focused on \textit{proper word choice}, and therefore excluded corrections of verb-tense, subject-verb agreement, word form, noun-number and article-or-determiner.  We included grammatical errors and lexical substitutions that correspond to improper word choice. Examples of some of the corrections suggested are presented in Figure \ref{fig:dataset}. Grammatical errors (according to the ConLL-2014 error dichotomy) include wrong collocation, and preposition and verb modal types of errors. Lexical substitution refers to substitutions that were preferred by the English editors. In cases in which more than one correction was made in a sentence, the sentence was adjusted so that it only requires the replacement of one word in the test stage.
The scientific-text test is available for use and may be downloaded by the research community.\footnote{\url{https://github.com/vicmak/Exploiting-BiLSTM-for-Proper-Word-Choice}}

\subsection{General Purpose Writing}

\noindent \textbf{General purpose English corpora.} For a representative and balanced English corpora we use the Corpus of Contemporary American English (COCA).\footnote{\url{http://corpus.byu.edu/coca/}} COCA is equally divided into five parts: spoken, fiction, popular magazines, newspapers, and academic texts, from the years 1990-2015, and has a total of 520 million words. The spoken part was excluded from our experiment. After limiting the vocabulary size to 30000 words and the sentence length to 40 words, a reduced corpus with 57 million words in 4.8 million sentences remained.

\noindent \textbf{General purpose text test set.} For a general purpose test set we use a subset of  the CoNLL-2014 GEC shared task dataset \cite{ng-EtAl:2014:W14-17}, that consists of 2421 sentences. Table \ref{table:conll} specifies the subset of errors that were chosen for the evaluation. The error distribution with the results is also presented in Table \ref{table:conll}.

As with the scientific writing, we only use those errors where a single word is replaced by another single word. However, as opposed to the scientific test set, we left all other grammatical errors in these sentences (in addition to the word choice errors), since we aimed to handle this in a manner similar to that which was done in the most recent study addressing the GEC problem \cite{rozovskaya2016grammatical}. 

\subsection{Human Annotators' Evaluation}

Scientific writing and general purpose evaluation scenarios underestimate the MRR, providing its lower bound. To further investigate our method's performance and approach to the actual MRR, we performed a human evaluation. 
In this evaluation scenario, we employed two professional English writers. Both of them are native English speakers, and obtained an undergraduate degree, one in the US and the other in the UK. Each annotator was given the list of 176 ESL writer's sentences with a target word and a correction for each sentence. A list of possible corrections for each sentence, generated by our method, was also given to each annotator. Each annotator was asked to select from among the possible corrections provided by our method, those corrections that would be an equal quality substitution for the original correction correction of the sentence made when the article was being edited prior to its submission to a conference or journal. Because the two annotators did not agree completely on the corrections, we collected all of the corrections that were selected by the annotators and created two comparison sets used to evaluate our method's performance:
\begin{enumerate}
\item \textbf{Combined set.} We combined all of the corrections selected by the two annotators, thereby obtaining a wider list of possible corrections.
\item \textbf{Intersection set.} We chose only the corrections that both annotators agreed on, thereby, receiving a more restricted, and likely more conservative correction list.
\end{enumerate}

We further evaluated our method with these two test sets.

\subsection{Implementation details}

 Due to the very large training sets and the extensive amount of time it took to train the models, we had to implement batch training. Batch training on RNNs when there are sentences of varying lengths in one batch is a very challenging engineering task. Therefore, we designed a special training mechanism that would take into account the two LSTM networks (right-to-left and left-to-right) and would not sum the errors obtained from padded sentences that are less then 40 tokens long. Thus, we implemented all RNN based models using the DyNet \cite{DyNet} toolkit\footnote{The implementation is available for download at: \url{https://github.com/vicmak/Exploiting-BiLSTM-for-Proper-Word-Choice}}. DyNet supports a dynamic computation graph, making this implementation possible. Standard and more common deep learning libraries, e.g., TensorFlow \cite{tensorflow2015-whitepaper}, do not support dynamic computation graphs, and enforce specifying the exact dimensions of input sentences. These implementations are provided with a list of sentences and ESL learner's original word explicitly specified in each sentence.

We used the NLTK \cite{bird06nltk} API for POS tagging in our two experimental strategies: 1) to access the WordNet database for words' senses and 2) to obtain statistical POS tagging.
We used the KENLM \cite {Heafield-estimate} toolkit\footnote{Available for download at: \url{https://github.com/kpu/kenlm}} to implement the n-gram LM with Kneser-Ney smoothing for the n-gram LM evaluation.

\subsection{Hyper-parameters}

We experimented with various hyper-parameter values and experienced a considerable trade-off between training time and quality. We were also limited by hardware resources, as we performed the training on an NVIDIA GRID K2 GPU with 4 GB of memory. The results reported here were obtained using the following hyper-parameters for the model architecture and training.

\begin{itemize}
\item Word embedding dimension = 200
\item Number of LSTM hidden units = 200
\item Sentential context embedding dimension = 400
\item Training batch size = 100 sentences
\item Maximal sentence length = 40 words for all RNN based methods (BiLSTM and RNNLM in Table \ref{table:scientif-results}) 
\item Vocabulary size = 30,000 (the size of initial one-hot vectors, representing the words as the starting point for word embedding learning)
\item N-gram LM with Kneser-Ney smoothing was trained to learn a distribution for five n-gram's sizes, starting from unigram to five-gram. 
\end{itemize}

In the training of RNN based models, we padded the sentences which were in lowercase letters with special \texttt{<start>} and \texttt{<stop>} tokens at the beginning and at the end of each sentence, to ensure consistent starting and ending point for each RNN's LSTM state.

Because the training batch size affects both training time and classification performance \cite{batch}, we experimented with both pure stochastic and mini-batch training procedures. The training on the scientific corpus took 24 hours to perform, with minibatch training and a batch size of 100 sentences. The time it took to train the general-purpose corpus (COCA) was 40 hours, also with minibatching and a batch size of 100 sentences. Stochastic training on the scientific corpus took took approximately 100 hours to train, with one sentence at a time with only one pass (epoch) over the data. The resulting performance was significantly poorer than that obtained in the case of minibatch training. 

%% file: 5_Results.tex
\newpage
\section{Results}
\label{section:results}

\begin{table}[h!]
\centering
\caption{Examples of the suggestions for the scientific writing test set. The errors are marked with an asterisk (*) and the correction follows the the forward slash (/). The top three correction suggestions are listed for each sentence.}

\begin{tabular}{|l|l|}
\hline
\textbf{Corrected sentence} & \textbf{Top 3} \\ \hline
\begin{tabular}[c]{@{}l@{}}The results clearly \textbf{*indicate/show} that our method \\ significantly  outperforms the current state-of-the-art.\end{tabular} & \begin{tabular}[c]{@{}l@{}}1) show\\ 2) demonstrate\\ 3) suggest\end{tabular} \\ \hline
\begin{tabular}[c]{@{}l@{}}Other predictors incorporate the statistical data of query term, \\ frequencies within the collection obtained \textbf{*from/during} the \\ indexing stage.\end{tabular} & \begin{tabular}[c]{@{}l@{}}1) at\\ 2) from\\ 3) during\end{tabular} \\ \hline
\begin{tabular}[c]{@{}l@{}}This technique mimics the number of operations \\ \textbf{*needed/required}  to transfer one text string to another.\end{tabular} & \begin{tabular}[c]{@{}l@{}}1) required\\ 2) needed\\ 3) used\end{tabular} \\ \hline
\begin{tabular}[c]{@{}l@{}}While the problem of product recommendations for consumers in\\  e-commerce, has been \textbf{*widely/extensively} analyzed modelling \\ consumer's  willingness, to pay for the recommended product has \\ hardly been addressed\end{tabular} & \begin{tabular}[c]{@{}l@{}}1) widely \\ 2) extensively \\ 3) previously\end{tabular} \\ \hline
\begin{tabular}[c]{@{}l@{}}In order to \textbf{*enforce/ensure} that out of budget features are not \\ selected,  the framework maintains a so called potential features set\end{tabular} & \begin{tabular}[c]{@{}l@{}}1) ensure \\ 2) guarantee \\ 3) find\end{tabular} \\ \hline
\end{tabular}
\label{table:scientific}
\end{table} 
\FloatBarrier

\subsection{Scientific Writing}

\begin{table}[]
\centering
\caption{Lower bound (underestimate) MRR for the scientific test set. The proposed bidirectional LSTM model trained on a domain specific corpus substantially outperforms multiple baselines which also employ only unlabeled data for constructing a statistical classifier.}
\begin{tabular}{|l|c|}
\hline
\multicolumn{1}{|c|}{\textbf{Model}} & \multicolumn{1}{l|}{\textbf{MRR}} \\ \hline
BiLSTM - Domain specific, batch train,  WordNet POS & 0.41 \\ \hline
BiLSTM - Domain specific, batch train, NLTK POS & 0.40 \\ \hline
BiLSTM - Domain specific, stochastic train, WordNet POS & 0.22 \\ \hline
BiLSTM - General purpose (COCA), batch train & 0.33 \\ \hline
N-Gram with Kneser Ney smoothing - Domain specific & 0.34 \\ \hline
Left-to-right RNNLM - Domain specific & 0.16 \\ \hline
Right-to-left RNNLM - Domain specific & 0.02 \\ \hline
context2vec t2c-t2t - General purpose & 0.11 \\ \hline
context2vec t2c - General purpose & 0.06 \\ \hline
\end{tabular}
\label{table:scientif-results}
\end{table}

\subsubsection{Baselines}

In order to demonstrate the strength of our bidirectional LSTM network model, we evaluated several baselines using the 176-sentences scientific test set:
\begin{enumerate}
\item A left-to-right RNNLM, trained on the high quality scientific corpus.
\item A right-to-left RNNLM, trained on the high quality scientific corpus.
\item A five-gram LM with Kneser-Ney smoothing, trained on the high quality scientific corpus. We derived a left-to-right five-gram LM. There are sentences in which the correction takes place within the first five words of a sentence. To accommodate for this, we use a whole sentence perplexity to rank the candidates. Thus, we take into account not only the history, but also the future, of the target word which is corrected.
\item Our bidirectional LSTM (BiLSTM) model trained on the general purpose COCA corpus. This baseline is used primarily to compare our model with a pre-trained \textit{general-purpose} context2vec model.
\item Our bidirectional LSTM (BiLSTM) model trained on the high quality scientific corpus, using NLTK POS tagger.
\item Our bidirectional LSTM (BiLSTM) model trained \textit{stochastically} on the high quality scientific corpus, using WordNet for POS tags. We use this benchmark to show the necessity of batch training.
\item A \textit{context2vec} t2c-t2t model, pre-trained on the UkWac \cite{Ferraresi08introducingand} general purpose corpus.
\item A \textit{context2vec} t2t model, pre-trained on the UkWac general purpose corpus.
\end{enumerate}

These baselines allow us to assess the impact of three different dimensions, on the performance of a proper word choice:
\begin{enumerate}
\item The model itself.
\item The POS filtering source.
\item The corpus of model training.
\end{enumerate}

\subsubsection{Improving the baselines}
Using the scientific test set, a high-quality academic corpus, and WordNet POS source, BiLSTM model achieves the highest result: an MRR of 0.41. In other words, the gold standard correction is routinely among the replacement candidates suggested by our method, falling, on average, between the second and the third suggested candidates. Table \ref{table:scientific} presents a few examples of corrected sentences from the scientific-text test set, along with the suggestion provided by our method. We compared our results to the baseline models. 
The result of MRR=0.41 only slightly outperforms the result obtained when using a standard, statistical POS tagger, which achieved an MRR of 0.40.
Surprisingly, a very respectable result was achieved by the well-known and strong benchmark of n-gram LM with Kneser-Ney smoothing. On average, its suggestions fell at the third place. Our BiLSTM model, which was trained stochastically, feeding one sentence at a time into the model during training, performed significantly worse, despite the fact that it used the same high quality scientific corpus. Moreover, BiLSTM model that was batch-trained on COCA corpus, performed even better than the model which was domain specific but stochastically trained.

Our BiLSTM model dramatically outperformed the uni-directional standard RNNLM models which achieved lower results in terms of word suggestion capabilities.

Finally, we compared our model, trained on COCA general purpose corpus, to a \textit{context2vec} model, which was pre-trained\footnote{Downloaded from: \url{http://u.cs.biu.ac.il/~nlp/resources/downloads/context2vec/}} on the two billion words ukWaC \cite{Ferraresi08introducingand} corpus. We experimented with two different context similarity measures that are available in \textit{context2vec}: 1) target-to-context ($t2c$) , and 2) combined similarity measure of target-to-context and target-to-target ($t2t$). This combined measure is denoted as $t2c-t2t$. Both $t2c$ and $t2t$  similarity measures are measured as a cosine distance between the word's embedding vector and the embedding vector of the context and word, respectively. In the $t2c-t2t$ similarity measure, the original target word is provided to the model, as it is provided to our suggestion method, and the combined similarity is measured as a product of the $t2c$ and $t2t$ measures. The general purpose COCA model outperforms the \textit{context2vec}'s model when the $t2c-t2t$ similarity measure or a simpler $t2c$ similarity measure is used. In addition, \textit{context2vec}'s model incorporates significantly more parameters than our model does. Specifically, the target words' embeddings and the sentential contexts' embeddings used by \textit{context2vec} in this setting  use a dimensionality of 600 units. Moreover, it uses 600 LSTM hidden and output units. \textit{context2vec}'s model includes an additional network - a multi layer perceptron (MLP), with non-linear activation functions, in order to capture complex context regularities. Therefore, this \textit{context2vec}'s model is significantly more complex and harder to train, again providing slightly worse results, when trained on another huge general purpose corpus.

The results for the scientific test set are summarized in Table \ref{table:scientif-results}.

\subsection{General-Purpose Writing}

\begin{table}[h!]
\centering
\caption{Lower bound (underestimate) MRR for the general-purpose test set. The MRR values for useful suggestions are indicated in \textbf{bold}.}
\begin{tabular}{|l|l|l|}
\hline
\textbf{Error type} & \textbf{\begin{tabular}[c]{@{}l@{}}Number of\\ sentences in the\\ test set\end{tabular}} & \textbf{MRR} \\ \hline
Verb tense & 205 & \textbf{0.31} \\ \hline
Verb modal & 56 & \textbf{0.41} \\ \hline
Verb form & 180 & 0.13 \\ \hline
\begin{tabular}[c]{@{}l@{}}Subject-verb\\ agreement\end{tabular} & 242 & \textbf{0.33} \\ \hline
Noun number & 399 & 0.15 \\ \hline
Pronoun form & 61 & \textbf{0.36} \\ \hline
Preposition & 447 & \textbf{0.41} \\ \hline
\begin{tabular}[c]{@{}l@{}}Wrong\\ collocation\end{tabular} & 547 & \textbf{0.20} \\ \hline
Word form & 158 & 0.06 \\ \hline
Parallelism & 17 & 0.15 \\ \hline
Linking words & 109 & \textbf{0.24} \\ \hline
\textbf{All errors} & \textbf{2421} & \textbf{0.25} \\ \hline
\end{tabular}

\label{table:conll}
\end{table}
\FloatBarrier
Table \ref{table:conll} contains the results for the subset of ConLL-2014 shared task test set on grammatical error correction. The best results, in which the gold standard correction was suggested by our method was between second and fifth place are marked in \textbf{bold}. The \textit{preposition}, \textit{pronoun form}, and \textit{verb modal} error types are handled especially well by our method, and the word replacement candidates suggested by our method often include the gold standard in second or third place. Remarkably, the method also performs quite well on the \textit{subject-verb agreement} error type. Of the suggestions provided by our method the gold standard correction comes in third place. 

\subsection{Human evaluation}

Thus far we computed the lower bound MRR, with a strict evaluation policy, where only one possible correction was used for the MRR computation. In the human evaluation scenario we use multiple possible corrections, that were obtained from two human annotators as described in Section \ref{section:evaluation}. As we expected, using a list of all possible corrections leads to better results as shown in Table \ref{table:human}.

\begin{table}[]
\centering
\caption{MRR results for human evaluated correct test sets. The combined correct list set, as expected, leads to the highest MRR value achieved by our method. The intersection set still leads to a high value. Both MRR values are higher than the lower bound MRR achieved by the BiLSTM model, compared to only one corrected option.}
\label{table:human}
\begin{tabular}{|l|l|}
\hline
\textbf{Correct list set} & \textbf{MRR} \\ \hline
Combined set & 0.61 \\ \hline
Intersection set & 0.49 \\ \hline
\end{tabular}
\end{table}

%% file: 6_Conclusions.tex
\section{Conclusions and Future Work}

We presented a simple and clear application of bidirectional LSTM tagger to the proper word choice task. Our implementation is aimed  at responding to the challenges faced by ESL scholars within academia, and their need to write in a highly professional and correct language in order to publish their studies. The system we have shown is simple, straightforward and applicable since it does not rely on manually annotated data. To emphasize the simplicity even further, our application does not require a training of a specific classifier for each type of grammatical error or lexical substitution, as opposed the the state-of-the-art GEC systems \cite{rozovskaya2016grammatical}. 

Our application addresses the types of errors  typical to ESL learners. Specifically, we  focus on the problem of word replacement, which may be due to wrong collocation correction or more intelligent lexical substitution.

In response to this problem, we propose a word choice suggestion method based on the application of bidirectional LSTM tagger. We demonstrated its performance in three evaluation settings. We performed the evaluation on a big subset of well known CoNLL-204 shared task (general purpose writing), and manually collected sentences (scientific writing) for further demonstration. We also contributed a new ESL learner test set to the NLP community that was collected from actual proofreading tasks and reflects a realistic scenario.  

In this paper we did not attempt to address the \textit{detection} of incorrect word usage, and therefore the results are not directly comparable to the existing GEC method, which handles detection and correction and uses the F-measure as its evaluation metric for the complete test set. In future research we intend to develop a word  detection method for detection poor words, in order to create a more comprehensive approach that can be compared with the state-of-the-art model presented by Rozovskaya and Roth \cite{rozovskaya2016grammatical}. We also intend to explore other similarity and substitutability measures proposed by Melamud et al. \cite{melamud2015simple}\\
One limitation of the method presented, that we will explore in future work,  is its ineffectiveness in handling  multiple interacting errors in one sentence.

%% file: 7_Appendix.tex
\appendix
\section{}

Just before the submission of the current paper for review, we decided to evaluate our method on its edited version.  We collected 15 sentences that contain the replacement of a word by another word and evaluated our method with these sentences. We achieved an MRR of 0.27 using the domain-specific (scientific) model and an MRR of 0.17 using the general-purpose (COCA) model. The top three suggestions for these sentences are presented in Table \ref{table:current}. 

\begin{table*}[h]
\centering
\caption{A sample of word replacements in the current paper, with the top three suggestions from domain-specific and general-purpose models. The replacement of the word marked with an asterisk (*) in a sentence was made by a professional English editor follows the forward flash (/).  }
\scriptsize
\begin{tabular}{|l|l|l|}
\hline
\multicolumn{1}{|c|}{\textbf{Sentence}} & \textbf{\begin{tabular}[c]{@{}l@{}}Domain specific \\ suggestions\end{tabular}} & \textbf{\begin{tabular}[c]{@{}l@{}}General purpose \\ suggestions\end{tabular}} \\ \hline
Scientific writing is *hard/difficult. & \begin{tabular}[c]{@{}l@{}}1) different\\ 2) important\\ 3) useful\end{tabular} & \begin{tabular}[c]{@{}l@{}}1) replete\\ 2) tantamount\\ 3) laughable\end{tabular} \\ \hline
\begin{tabular}[c]{@{}l@{}}A language model (LM) is defined as a \\ function that *puts/assigns  a probability \\ distribution to words in a given vocabulary.\end{tabular} & \begin{tabular}[c]{@{}l@{}}1) assigns\\ 2) maps\\ 3) associates\end{tabular} & \begin{tabular}[c]{@{}l@{}}1) is \\ 2) produces\\ 3) involves\end{tabular} \\ \hline
\begin{tabular}[c]{@{}l@{}}From this distribution, filter the words \\ *which/that  are tagged with  the same part of \\ speech  if replaced with the original word in \\ the sentence. \end{tabular} & \begin{tabular}[c]{@{}l@{}}1) that \\ 2) words\\ 3) phrases\end{tabular} & \begin{tabular}[c]{@{}l@{}}1) they\\ 2) and\\ 3) you\end{tabular} \\ \hline
\begin{tabular}[c]{@{}l@{}}Consider the *next/following sentence \\ which was  written by an ESL learner. \end{tabular} & \begin{tabular}[c]{@{}l@{}}1) following\\ 2) first \\ 3) next\end{tabular} & \begin{tabular}[c]{@{}l@{}}1) same\\ 2) last\\ 3) final\end{tabular} \\ \hline
\end{tabular}

\label{table:current}
\end{table*}